\documentclass{article}

\usepackage{microtype}
\usepackage{graphicx}
\usepackage{subfigure}
\usepackage{booktabs} 
\usepackage{amsmath}
\usepackage{hyperref}
\usepackage{flexisym}
\usepackage{float}
\usepackage{mathtools}
\usepackage{multirow}
\usepackage{xcolor}
\usepackage{algpseudocode}

\graphicspath{./images/}

\newcommand{\ours}{MEDIC}

\usepackage[accepted]{icml2018_arxiv}

\newcommand{\papertitle}{Incremental Learning with Maximum Entropy Regularization:\\ Rethinking Forgetting and Intransigence} 
\icmltitlerunning{\papertitle}

\begin{document}

\twocolumn[
\icmltitle{\papertitle}



\icmlsetsymbol{equal}{*}

\begin{icmlauthorlist}
\icmlauthor{Dahyun Kim}{gist}
\icmlauthor{Jihwan Bae}{gist}
\icmlauthor{Yeonsik Jo}{gist}
\icmlauthor{Jonghyun Choi}{gist}
\end{icmlauthorlist}

\icmlaffiliation{gist}{School of Electrical Engineering and Computer Science, Gwangju Institute of Science and Technology (GIST), Gwangju, South Korea}

\icmlcorrespondingauthor{Jonghyun Choi}{jhc@gist.ac.kr}

\icmlkeywords{Machine Learning, ICML}

\vskip 0.3in
]



\printAffiliationsAndNotice{}  

\begin{abstract}
Incremental learning suffers from two challenging problems; forgetting of old knowledge and intransigence on learning new knowledge. 
Prediction by the model incrementally learned with a subset of the dataset are thus uncertain and the uncertainty accumulates through the tasks by knowledge transfer. To prevent overfitting to the uncertain knowledge, we propose to penalize confident fitting to the uncertain knowledge by the Maximum Entropy Regularizer (MER). Additionally, to reduce class imbalance and induce a self-paced curriculum on new classes, we exclude a few samples from the new classes in every mini-batch, which we call \emph{DropOut Sampling (DOS)}.
We further rethink evaluation metrics for forgetting and intransigence in incremental learning by tracking each sample's confusion at the transition of a task since the existing metrics that compute the difference in accuracy are often misleading.
We show that the proposed method, named \emph{`MEDIC'}, outperforms the state-of-the-art incremental learning algorithms in accuracy, forgetting, and intransigence measured by both the existing and the proposed metrics by a large margin in extensive empirical validations on CIFAR100 and a popular subset of ImageNet dataset (TinyImageNet).
\end{abstract}

\section{Introduction}

Incremental learning is a paradigm in which the model is expected to learn a set of tasks sequentially, where the task is a machine learning problem such as classification of a set of classes. By the success of deep neural networks in machine learning~\cite{krizhevskySH12, DBLP:journals/corr/abs-1810-04805, SilverHuangEtAl16nature}, there are many proposals for incremental learning using deep neural networks~\cite{Rebuffi_2017_CVPR, Kirkpatrick2017OvercomingCF, castro2018eccv, Chaudhry_2018_ECCV}. 
The neural networks, however, have been known to suffer from two prominent problems especially in the incremental learning set-up; catastrophic forgetting of the old knowledge and intransigence on learning new knowledge. 
While the forgetting may be less harmful in the beginning of the incremental learning process (in the early tasks), the knowledge on past tasks vanishes as the learning progress.
In addition, since the budget for storing the samples from the old group of classes (\emph{e.g.,} previously given groups of classes in classification), is limited for practicality in incremental learning, there are a larger number of samples belonging to the new group of classes than the ones belonging to the old group of classes. The imbalance of the old and new classes makes the model prone to overfitting to the new task by which worsens forgetting. On the other hand, attempting to overcome forgetting by giving extra supervision on old classes might hinder the model from learning new classes, causing intransigence. 

To alleviate both problems, we propose to regularize a model to optimize less on the uncertain transferred-knowledge by using the maximum entropy regularizer (MER)~\cite{DBLP:journals/corr/PereyraTCKH17, jaynes57} and exclude a number of samples in the new group of classes during stochastic gradient descent of a mini-batch (which we call \emph{DropOut Sampling (DOS)}. 
The MER forces the model to less fit to the uncertain transferred-knowledge, avoiding optimizing on potentially \emph{incorrect} knowledge.
Inspired by curriculum learning \cite{Bengio2009CL} and self-paced learning \cite{Kumar2010SelfPaced} literature, the DOS induces a curriculum for the new class samples in a self-paced manner, thus allowing our model to achieve better intransigence on learning the new classes. Furthermore, the selective removal of samples from the new classes also leads to less forgetting of the old classes because the class imbalance is reduced.
We name our method as \textbf{`MEDIC'} (\emph{ME}R and \emph{D}OS for \emph{I}n\emph{C}remental learning) that cures both forgetting and intransigence in incremental learning.

In evaluating incremental learning algorithms, the classification accuarcy at the final task has been widely used ~\cite{castro2018eccv, Rebuffi_2017_CVPR, Liu2018RotateYN}. However, the accuracy does not measure the resilence of a model to catastrophic forgetting (forgetting) or the inability to learn new groups of classes (intransigence). 
Recently, F and I measures are proposed for measuring forgetting and intransigence by calculating difference of overall classification accuracy across the tasks~\cite{Chaudhry_2018_ECCV}.
They, however, do not encode relative significance of the difference between task accuracies in different magnitudes. 
For instance, a difference of 0.1 between 0.9 and 0.8 has different implication than a difference of 0.1 between 0.5 and 0.4. 
Thus, we propose to encode relative significance of the difference in a new measure.
More importantly, the metrics may interprete behaviors other than forgetting as forgetting due to confusing classes. 
Please refer to Sec.~\ref{sec:newmetric} for a detailed argument with examples.

To address the issues of the existing evaluation metrics, we propose a new metric called \emph{Sample Dynamics} (SD). 
The SD tracks sample behavior of the reference model which is trained with all the training data for all the classes present in a given task and an incremental model. 
It allows us to consider the samples that does not confuse the reference model, \emph{i.e,} samples that are not in the confusable classes. In addition, the SD is defined as a ratio instead of a difference, allowing the measurement to be invaraint to different magnitudes of the classification performance.

We show that the MER and the DOS improve performance of the final classification accuracy, the F and the I proposed by~\cite{Chaudhry_2018_ECCV}, and the proposed SDI and SDF by an ablation study (Sec.~\ref{sec:abal}).
We also show that the proposed `MEDIC' significantly outperforms the state-of-the-art incremental learning methods by a large margin (more than 5\% in accuracy) on two image classification datasets; the CIFAR100 and the TinyImageNet dataset.
We further analyze the methods in two extreme incremental task configurations, the best-first and the worst-first, and find that the proposed method outperforms in any scenario.
Also in a budgeted incremental setup where the budget for the memory of old classes' samples decreases, we show that the performance drop by the `MEDIC' is the least among all methods.

\section{Related Work}

\paragraph{Catastrophic Forgetting.}
Catastrophic interference or catastrophic forgetting has been a well-known problem of neural networks~\cite{mccloskeyC89, ratcliff90}.
iSince the pioneering work of \cite{Li2017LearningWF} tackling catastrophic forgetting in incremental learning using knowledge distillation~\cite{hintondistillation}, there are a number of proposals dealing with the forgetting. Lee \emph{et al.} and Belouadah \emph{et al.} propose a network architecture whose capacity is dynamically expandable~\cite{Lee2017LifelongLW, DBLP:journals/corr/abs-1808-06396}.
Other strategies include mitigating catastrophic forgetting by introducing dual-memory system and calculating the importance of the parameters of a neural network~\cite{DBLP:journals/corr/abs-1711-10563, Aljundi2018MemoryAS}, or learning a mask in terms of network quantization and pruning network parameters~\cite{DBLP:journals/corr/abs-1711-05769, Mallya2018PiggybackAA, pmlr-v80-serra18a}. However, they are not scalable since the network size grows proportionally to the size of tasks. 

Kirkpatrick \emph{et al.} propose Elastic Weight Consolidation (EWC) to constrain important network parameters (weights) to be in the vicinity of their outdated values by using the posterior of the network parameters given the dataset and the Fisher Information Matrix (FIM)~\cite{Kirkpatrick2017OvercomingCF}. Recently, \cite{Liu2018RotateYN} proposed R-EWC to overcome the assumption that FIM should be diagonal by rotating the parameter space of the neural network while preserving the output of the forward pass. 
As R-EWC adds two additional convolutional layers for the rotation, they increase the network capacity. This makes direct comparison between R-EWC and \ours\ less meaningful as the network capacity is fixed in \ours. \cite{Chaudhry_2018_ECCV} propose a generalized form of EWC by the empricial FIM as an approximation for the KL-Divergence between the network parameters of the previous task and the ones of a new task. They also introduce a parameter importance score that is similar to \cite{Zenke2017ContinualLT} to capture the parameter's importance relative to the optimization path. Furthermore, they introduce metrics for incremental learning algorithms that indicate how well an algorithm overcomes catastrophic forgetting as well as intransigence, an ability to learn a new group of classes. 
Although the contributions of R-Walk are well thought of, the gain over EWC is small (0.4\%) and hence we decided to compare with the original EWC. 

\paragraph{Maximum Entropy Regularization (MER).}
The maximum entropy principle or regularization (MER) is proposed to prevent overfitting by enforcing the exploration in predicted labels~\cite{jaynes57}.
It has been widely used in unsupervised learning~\cite{Tyers2015UnsupervisedTO}, reinforcement learning~\cite{DBLP:journals/corr/MnihBMGLHSK16, DBLP:journals/corr/LuoCJS16, DBLP:journals/corr/NorouziBCJSWS16} and in the supervised settting~\cite{DBLP:journals/corr/PereyraTCKH17}.
Interestingly, Szegedy \emph{et al.} claim that label smoothing achieves better generalization by similar reason of the MER~\cite{szegedyLJSRAEVR15}.
The work using the MER in the most similar problem setup to the incremental learning is \cite{NIPS2018_7363}. 
They use the MER for the sequential prediction or sequence learning problem which does not require to remember the knowledge learned in the previous tasks.
To our best knowledge, we are the first in employing the MER for the incremental learning which is a more challenging set-up than the sequence learning since it requires the model to remember old tasks and learn new tasks continuously.

\paragraph{Knowledge Distillation.}
One of the popular methods to transfer the knowledge from an old task (teacher) to the new one (student) is knowledge distillation (KD)~\cite{hintondistillation}. Li \emph{et al.} use it for supervised learning by formulating a loss function as a sum of the standard cross-entropy and the distilliation loss~\cite{Li2017LearningWF}.
However, they assume that the old tasks and new tasks originate from different domain.
Castro \emph{et al.} propose a KD method without the assumption as to transfer knowledge from the new and old taskss that can come from a related distribution~\cite{castro2018eccv}. 


Interestingly, Furlanello \emph{et al.} show that network trained with the distilled knowledge outperforms the original network by training the student network in an iterative manner~\cite{Furlanello2018BornAN}. 
Although this setup looks similar to the incremental learning in terms of iterative training of model, it is on a fixed task while the incremental learning is on a gradually increasing set of tasks. 



\section{Approach}


The objective function for incremental learning consists of two terms, the learning objective and the knowledge transfer (from past tasks to a new task) objective, and can be simply written as follows:
\begin{equation}
    L = L_{learn} + \alpha L_{transfer},
\label{eq:original}
\end{equation}
where $\alpha$ is a balancing hyper-parameter. Since we deal with classification, we use cross entropy ($CE$) between the ground-truth labels $p$ and the predicted labels $q$ for $L_{learn}$, thus $L_{learn} = CE(p,q)$.
For transferring knowledge from previous tasks (group of classes to classify) to a new task, knowledge distillation~\cite{hintondistillation} is one of the widely used methods~\cite{Li2017LearningWF, castro2018eccv}.
We also use the knowledge distillation loss $L_D$ for the $L_{transfer}$. 
Note that the $L_D$ at task $T$ is accumulated as training goes on, hence the $L_D$ is given as the summations of $L_D$ over past tasks up to $T-1$. We can rewrite the objective function as:
\begin{equation}
    L(p,q) = CE(p,q) + \alpha \displaystyle \sum_{k=1}^{T-1} L_{D}(k, p^k,q),
\label{eq:main}
\end{equation}
where $L_{D}(k, p^k, q)$ is the distillation loss calculated on the classes belonging to task indexed with $k$ using the soft labels produced by the model a trained up to task $k-1$ denoted by $p^k$. 

For the distillation loss, cross entropy (CE) is again, one of the most popular choices~\cite{Li2017LearningWF, castro2018eccv}. 
While the CE can capture the difference between the soft labels of the past model (the teacher) and the softmax probabilities of the current model (the student), it becomes less helpful when the teacher's soft labels are uncertain because the student would try to fit to the wrong soft labels, which is often the case in incremental learning.
Thus, when using distillation in incremental learning, it is important to account for the uncertainty in the given soft labels. 
One way of dealing with uncertain knowledge in distillation is to make the teacher's soft labels smoother by dividing the logits with a temperature value that is greater than 1~\cite{hintondistillation}. 
However, it has two drawbacks; first, when teacher's soft label's distribution is peaked (having high Kurtosis), the smoothing effect is not prominent.
Second, as the smoothing is done in the teacher's label, the uncertainty of the student's distribution is not considered.

\paragraph{Incremental Learning by Knowledge Distillation with the MER.} 
We propose to regularize the loss by enforcing the predictions to be less certain when dealing with uncertain transferred knowledge by the use of maximum entropy regularizer (MER)~\cite{DBLP:journals/corr/PereyraTCKH17}.
The regularizer lets the model transfer the knowledge from the past model in a conservative way as our model does not respond radically to high probabilities in the teacher's soft labels as much as the standard cross entropy does.
We formulate the distillation with the MER as:
\begin{equation}
\begin{aligned}
    L_{D}(k, p^k, q) &= CE(p^k, q) - H(q)\\
                  &= \frac{1}{N}\displaystyle \sum_{i=1}^{N} \displaystyle \sum_{j=1}^{C} (q_{ij} - p^k_{ij}) \log q_{ij},
\end{aligned}
\label{eq:ldl}
\end{equation}
where $q_{ij}$ are the softmax probabilities with $i$ being a sample index and $j$ being a class index, $H(q)$ is the entropy of the probability distribution $q$, the predicted label, $p^k_{ij}$ is the soft labels obtained by the model trained up to task $k$ and $N$, $C$ are the number of samples and number of old classes, respectively. 
We show the benefit of the maximum entropy regualarizer by empirical validations in Section~\ref{sec:abal}.

\paragraph{DropOut Sampling (DOS).}
Supervision to remember old classes disturbs learning new classes and can result in poor intransigence. 
In order to learn a model with both less forgetting and less intransigence, we propose a \emph{DropOut Sampling} that is performed at every mini-batch of the stochastic gradient descent (SGD) process, inspired by curriculum learning \cite{Bengio2009CL} and self-paced learning \cite{Kumar2010SelfPaced} literature.

For each mini-batch, we randomly exclude samples from the new classes at the early epoch of learning.
The removal of new class samples in the mini-batch alleviates the class imbalance, which would result in less forgetting. 
It can be viewed as a bootstrapping aggregation or bagging in a self-paced way \cite{Kumar2010SelfPaced} as it is done in mini-batch wise. 
Further, we employ the idea of curriculum learning~\cite{Bengio2009CL, AAAI159750} to further improve the ensembling effect.
In the later epoch of learning, the model's predictions become more certain. 
We exclude samples from the new classes with high CE value, inducing a self-paced curriculum in which samples with high uncertainty are learned later.
We describe the details of the algorithm in Algorithm~\ref{algo:dsf}.

\begin{algorithm}[t]
\caption{DropOut Sampling (DOS)}
\label{algo:dsf}
\begin{algorithmic}[1]
    \Function{DOS}{$H$, $e$} \Comment{For Mini-batch $H$ at Epoch $e$}
    \State $A = \{x | x \in \text{Old Classes}\}$
    \State $B = \{x | x \in \text{New Classes}\}$ \Comment{$H = A \cup B$}
    \State $\alpha = \max(\min(|A|, |B|), \frac{|B|}{2})$
    \If{$e \leq K$}
        \State $C = \textproc{SampleTopDescend}(B, Random, \alpha)$
    \Else
        \State $C = \textproc{SampleTopDescend}(B, CE, \alpha)$
    \EndIf
    \State \textbf{return} $H \setminus C$
    \EndFunction
    \Function{SampleTopDescend}{$S$, $F$, $\gamma$}
        \State $S' = \textproc{Sort}(S, F, \text{Descend})$ \Comment{Sort $S$ by a measure $F$ in descending order}
        \State \textbf{return} $\{ x_i | x_i \in S' \cap 0 \leq i < \gamma \}$
    \EndFunction
\end{algorithmic}
\end{algorithm}

\paragraph{Stochastic Sampling.}
When a new task is given, samples from old tasks are subsampled to constitute the subset of samples from old tasks. 
There are several sampling strategies proposed including mean-of-feature selection (MoF) \cite{Rebuffi_2017_CVPR, Chaudhry_2018_ECCV}, and herding selection \cite{castro2018eccv}. Since all proposals improves the accuracy but marginally, we choose a random selection scheme for the computational efficiency without much loss of performance for all our experiments.

\paragraph{Balanced Fine-Tuning.}
We also use the balanced fine-tuning proposed by \cite{castro2018eccv} that uses a balanced subset of samples in order to deal with the class imbalance between old group of classes and new group of classes.

\subsection{New Metric on Forgetting and Intransigence}
\label{sec:newmetric}

                                                                                                                                                                                                                                                                                                                                                                                                                                                                                                                                                                                                                                                                                                                                                                                                                                                                                                                                                                                                                                                                                                                                                                                                                                                                                                                                                                                                                                                                                                                                                                                                                                                                                                                                                                                                                                                                                                                                                                                                                                                                                                                                                                                                                                                                                                                                                                                                                                                                                                                                                                                                                                                                                                                                                                                                                                                                                                                                                                                                                                                                                                                                                                                                                                                                                                                                                                                                                                                                                                                                                                                                                                                                                                                                                                                                                                                                                                                                                                                                                                                                                                                                                                                                                                                                                                                                                                                                                                                                                                                                                                                                                                                                                                                                                                                                                                                                                                                                                                                                                                                                                                                                                                                                                                                                                                 Despite the importance of the catastrophic forgetting and intransigence in the incremental learning, numerous works~\cite{castro2018eccv,Rebuffi_2017_CVPR, Kirkpatrick2017OvercomingCF} have evaluated their algorithms only on classification accuracy.
Although, the accuracy may give an idea of the classifier's overall performance, it does not directly show an algorithm's resilliance to catastrophic forgetting and/or intransigence.

Recently, Chaudhry \emph{et al.} propose metrics other than classification accuracy to evaluate an incremental learning algorithm: F (forgetting) and I (intransigence)~\cite{Chaudhry_2018_ECCV}. 
The F measures the resilience of a model to forgetting by computing an average of the differences between the maximum task accuracy and the current task accuracy for the past tasks.
The I measures the inability to learn new groups of classes by the difference between the accuracy of a reference model on the current task and the accuracy of the incremental model of the current task.

However these metrics have two limitations. First, both the F and the I are formulated as differences across tasks. The difference, however, is not the best way to measure the amount of forgetting and intransigence. For example, if the accuracy drops by 10\% from 90\% (to 80\%) then the 11.1\% of original knowledge has been lost while the same drop of 10\% from 30\% to 20\% implies 33.3\% of original knowledge has been lost. In both scenarios, the F and I measure it as 10\% loss but is not appropriate.
Second, their forgetting measure may associate simple confusion amongst confusing classes as forgetting when it should not. Suppose that you are dealing with a dataset with confusing classes A and B. Then, in the first task, only class A is present and there is no confusion towards class B. When the second task is given, class B is in the new task and the accuracy on class A plummets, due to severe confusion with class B. 
Such accuracy drops should not be considered as forgetting because the confusion is caused by the confusable classes \emph{not} by forgetting. However, the F measure accounts this behavior as forgetting because the accuracy difference makes no distinction between genuine forgetting and simple confusion because.
To address the above issues, we revise the notion of forgetting and intransigience by tracking each sample's confusion when there comes new classes, in a ratio form. 
We call it as \emph{Sample Dynamics}.

\paragraph{Sample Dynamics.} 
We define two metrics for forgetting and intransigence by tracking how a sample's confusion changes at the transition from old tasks to a new one, and name them as Sample Dynamics-Forgetting (SDF) and Sample Dynamics-Instransigence (SDI), respectively.
Both metrics are computed by taking a ratio of how a sample's confusion is distributed amongst old or new group of classes when inferenced by an incremental model and by the reference model.

Specifically, let us denote the incrementally learned model that has been trained up to task $k$ as $M_k(\cdot)$, and the reference model for task $k$ as $R_k(\cdot)$. We divide the classes in task $j(\leq k)$ into \emph{old classes} (denoted as $D^{old}_j$) if the classes are in the old task, and \emph{new classes} (denoted as $D^{new}_{j}$), otherwise. 
Let $x_i$ be a sample in the dataset, $y_i$ be a corresponding ground truth label, $\hat{y}^{M_k}_i = M_k(x_i)$ be a predicted label of $x_i$ by $M_k(\cdot)$ and $\hat{y}^{R_k}_i = R_k(x_i)$ be a predicted label of $x_i$ by $R_k(\cdot)$.
We can write the forgetting of $M_k(\cdot)$ on task $j$, $f_{k,j}$, as the following:
\begin{equation}
\resizebox{0.90\linewidth}{!}{
    $f_{k,j} = \frac{\left| \{y_i |y_i \in D^{old}_j~\land~\hat{y}^{R_k}_i \in D^{old}_j~\land~\hat{y}^{M_k}_i \in D^{new}_j\} \right|}
                {\left| \{y_i | y_i \in D^{old}_j~\land~\hat{y}^{R_k}_i \in D^{old}_j\} \right|},$
}
\end{equation}
where $|\cdot|$ is the cardinality operator. 
$f_{k,j}$ is a ratio of the number of the samples that are in $D^{old}_j$ and the predictions made by $R_k$ are also in $D^{old}_j$ against number of the samples in $D^{old}_j$ and the predictions made by $R_k(\cdot)$ are also in $D^{old}_j$ but the predictions made by $M_k(\cdot)$ is in $D^{new}_j$. 

With the $f_{k,j}$ defined, we define the `Sample Dynamics - Forgetting' (SDF) for the $M_k$ as:
\begin{equation}
    SDF_k = \frac{1}{k} \sum_{j=2}^{k} f_{k,j}.
\end{equation}
Likewise, we denote the intransigence of $M_k(\cdot)$ on task $j$, $i_{k,j}$, as the following.
\begin{equation}
\resizebox{0.90\linewidth}{!}{
    $i_{k,j} = \frac{\left| \{y_i |y_i \in D^{new}_j~\land~\hat{y}^{R_k}_i \in D^{new}_j~\land~\hat{y}^{M_k}_i \in D^{old}_j\} \right|}
                {\left| \{y_i | y_i \in D^{new}_j~\land~\hat{y}^{R_k}_i \in D^{new}_j\} \right|}.$
}
\end{equation}
It is a ratio of number of samples in $D^{new}_j$ and the predictions made by $R_k(\cdot)$ are also in $D^{new}_j$ against the number of samples in $D^{new}_j$ and the predictions made by $R_k(\cdot)$ is in $D^{new}_j$ but the predictions made by $M_k(\cdot)$ is in $D^{old}_j$. 

With the $i_{k,j}$ defined, we define the `Sample Dynamics - Intransigience' (SDI) for the $M_k$ as:
\begin{equation}
    SDI_k = \frac{1}{k} \displaystyle \sum_{j=2}^{k} i_{k,j}.
\end{equation} 

\noindent \textit{Averaged Sample Dynamics.} 
Both the $SDF_k$ and the $SDI_k$ are the measure at the end of $k^{\text{th}}$ step of incremental learning, thus they capture less on how the forgetting and intransigence changes in the intermediate learning steps.
We further define $SDI_{avg}$ and $SDF_{avg}$ by averaging the $SDI_k$ and $SDF_k$ throughout the tasks.
In situations where the last set of classes have considerable differences with the past set of classes, $SDF_{avg}$ and ${SDI}_{avg}$ provide useful insights as described in Section \ref{sec:sota}.


\section{Experiments}
\label{sec:exp}

\begin{table*}[t!]
\centering
\resizebox{0.77\paperwidth}{!}{
\begin{tabular}{c|c|c|c|c||c|c|c|c}
\hline
Unit(\%) & Accuracy $\uparrow$ & A10 $\uparrow$ & F10 $\downarrow$ & I10 $\downarrow$ & SDF10 $\downarrow$ & SDI10 $\downarrow$ & $\text{SDF}_{avg}$ $\downarrow$ & $\text{SDI}_{avg}$ $\downarrow$ \\ \hline
Baseline & 62.64$\pm$0.26 &  64.40$\pm$0.29 &  0.21$\pm$0.32 &  25.01$\pm$0.33 &  10.17$\pm$0.36 &  18.80$\pm$0.31 &  6.49$\pm$0.10 &  8.42$\pm$0.11 \\ \hline \hline
\ours\ w/o MER, DOS & 68.32$\pm$0.09 &  67.16$\pm$0.42 &  1.59$\pm$0.37 &  16.03$\pm$0.28 &  5.96$\pm$0.27 &  20.25$\pm$0.46 &  3.72$\pm$0.11 &  9.69$\pm$0.13 \\ \hline
\ours\ w/o DOS & 70.07$\pm$0.19 &  70.42$\pm$0.07 &  0.31$\pm$0.16 &  12.78$\pm$0.18 &  5.51$\pm$0.18 &  16.61$\pm$0.55 &  3.17$\pm$0.05 &  8.55$\pm$0.13 \\ \hline
\ours\ w/o MER & 69.65$\pm$0.18 &  70.41$\pm$0.21 &  \textbf{-0.14$\pm$0.35} &  13.29$\pm$0.15 &  5.38$\pm$0.05 &  16.71$\pm$0.46 &  3.56$\pm$0.08 &  8.19$\pm$0.12 \\ \hline \hline
\ours\ (Full Model) & \textbf{72.51$\pm$0.17} &  \textbf{73.35$\pm$0.35} &  0.13$\pm$0.16 &  \textbf{9.68$\pm$0.44} &  \textbf{5.05$\pm$0.10} &  \textbf{13.14$\pm$0.28} &  \textbf{2.96$\pm$0.06} &  \textbf{6.55$\pm$0.26}
 \\ \hline
\end{tabular}
}
\caption{\textbf{Ablation Study.} $\uparrow$ indicates higher number is better, $\downarrow$ otherwise. Accuracy is averaged classification accuracy from task 1 to task 10. Numbers after $\pm$ denotes standard deviation. Results on the eight metrics for methods tested on the Random configuration using the CIFAR100 dataset. Best results are in bold for each of the eight metrics.}
\vspace{-1em}
\label{table:ablation}
\end{table*}

\subsection{Experimental Set-up}

\paragraph{Datasets.} 
We use CIFAR 100~\cite{Krizhevsky09learningmultiple} and TinyImageNet ~\cite{TinyImageNet} for our experimental validations. 
The CIFAR100 consists of 60,000 images with the size of 32$\times$32 pixels in 100 object categories. Among 60,000, 50,000 images are used for training a model and 10,000 images are used for testing. 
The TinyImageNet consists of 120,000 images with size of 64 $\times$ 64 pixels in 200 object categories. Of the 120,000 images, 100,000 are used for training, 10,000 for validation, and 10,000 for testing. 
Note that we opted to use the validation set when testing instead of submitting predicted labels on the testset to the evaluation server held by Standford. 

Data augmentations such as contrast, mirroring, flipping, and normalization are done on the CIFAR100 dataset to achieve better generalization.

\paragraph{Model Architectures.}
We use the DenseNet121 \cite{Huang2017DenselyCC} as a backbone for all our experiments, with minor changes for the input image size. 

\paragraph{Memory for the Samples from Old Classes.}
For the size of memory for the old group of classes, we chose 2,000 for most of experiments on CIFAR100 following \cite{Rebuffi_2017_CVPR}, except for the experiments in Sec.~\ref{sec:budget1000} where we chose 1000. 
We chose 4,000 for all experiments on TinyImageNet. 




\paragraph{Task Definition.}
For CIFAR100 experiments, we split the classes into 10 groups of 10 classes and follow a standard task definition following \cite{castro2018eccv, Rebuffi_2017_CVPR}, and 10 groups of 20 classes for the TinyImageNet. 
The standard task definition is that the $k^{\text{th}}$ task is composed of $10 \times (k-1)$ old classes and 10 new classes for CIFAR100 and $20 \times (k-1)$ old classes and 20 new classes for TinyImageNet. We test the model on all the classes it has seen so far, \emph{i.e} $10 \cdot k$ for CIFAR100 and $20 \times k$ for TinyImageNet. 
Note that \cite{Chaudhry_2018_ECCV} do not follow the standard task definition but define a task as classifying 10 new classes at each step of increment but allows the model's confusion to span over all $10 \times k$ classes. Therefore the results reported in \cite{Chaudhry_2018_ECCV} are not directly comparable to our results.

\paragraph{Task Configuration.}
We define the order of group of classes given to the model as \emph{`task configuration'}. The task configuration should be blind to the incremental learning algorithm. As such, we conduct experiments on three random task configurations same as the previous work~\cite{Rebuffi_2017_CVPR, castro2018eccv, Chaudhry_2018_ECCV, Kirkpatrick2017OvercomingCF} unless otherwise mentioned.

On the other hand, because it is expected that the performance of incremental learning highly depends on configurations of given task, \emph{i.e.,} order of set of classes given, we also analyze incremental learning methods in two extreme task configurations; best first and worst first for evaluation purposes.
The best first and worst first configurations are obtained by sorting the classes by the classwise accuracy of the reference model.
 


\paragraph{Evaluation Metrics.}
For comprehensive evaluations, we use metrics proposed in the literature and the proposed metrics. The metrics include average classification accuracy on task 1-10 (Accuracy), the three measures proposed in~\cite{Chaudhry_2018_ECCV} (average task accuracy (A10), forgetting measure (F10), intransigence (I10) on the model trained up to task 10), and four of the proposed measures (SDF10, SDI10 on the model trained up to task 10, $\text{SDF}_{avg}$ and $\text{SDI}_{avg}$). 

\subsection{Ablation Study}
\label{sec:abal}

\begin{figure}[t!]
\centering
\includegraphics[width=1\columnwidth]{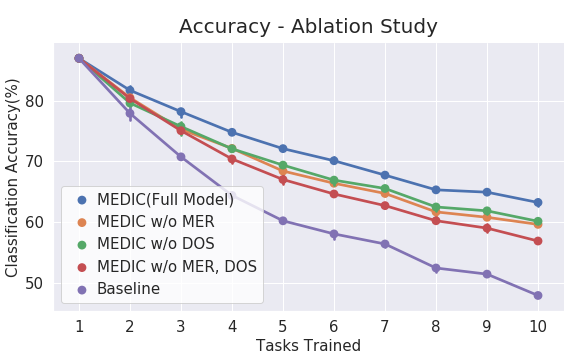}\\
\includegraphics[width=1\columnwidth]{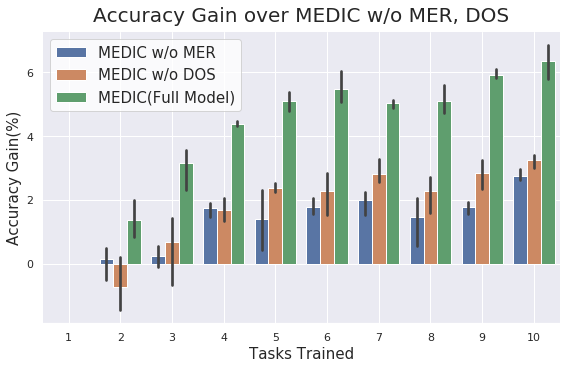}\\
\vspace{-3mm}
\caption{\textbf{Classification Accuracy in the Ablation Study.} \textbf{(Top)} Accuracy comparison. The y-axis indicates the classification accuracy and the x-axis indicates the task number. \textbf{(Bottom)} Accuracy gain over \ours\ without the MER and the DOS for variants of \ours. The y-axis indicates the gain in accuracy over \ours\ without the MER and the DOS. Both plots are shown with error bars from three trials. Refer to Section \ref{sec:abal} for details.} 
\vspace{-3mm}
\label{fig:ablation}
\end{figure}

We investigate the contributions of each component of our model; the Maximum Entropy Regularization (MER) and DropOut Sampling (DOS) and summarize the results in Table \ref{table:ablation}. 
The baseline refers to the DenseNet + Sample Memory and others are ablations of our model.

The MER helps the model to less fit to the uncertain soft labels, leading to better forgetting and intransigence.
The DOS reduces class imbalance and induces a self-paced curriculum on new classes, improving both forgetting and intransigence.

Further, the benefit by the MER and the DOS generally increases as the tasks progress as shown in Figure~\ref{fig:ablation} (Bottom). 
It is expected because the uncertainty of the transferred knowledge is accumulated and the MER and DOS would prevent the overfitting to the uncertain knowledge that is propagated as the learning step progresses.

Particularly at the early tasks, the MER may harm the classification accuarcy  because the soft labels at the very early step of the incremental learning might be certain enough for the MER to lose its benefits (See the dark orange bar in the task 2 in Figure~\ref{fig:ablation} (Bottom)). However, we can see benefit of MER in all tasks other than the task 2.

\subsection{Comparison with State of the Arts}
\label{sec:sota}
\begin{table*}[!th]
\centering
\resizebox{0.77\paperwidth}{!}{
\begin{tabular}{c|c|c|c|c|c||c|c|c|c}
\hline
      & Unit(\%) & Accuracy $\uparrow$ & A10 $\uparrow$ & F10 $\downarrow$ & I10 $\downarrow$ & SDF10 $\downarrow$ & SDI10 $\downarrow$ & $\text{SDF}_{avg}$ $\downarrow$ & $\text{SDI}_{avg}$ $\downarrow$ \\ \hline\hline
      
    \parbox[t]{2mm}{\multirow{4}{*}{\rotatebox[origin=c]{90}{Rand. Cfg}}} 
      & \ours & \textbf{72.51$\pm$0.17} &  \textbf{73.35$\pm$0.35} &  \textbf{0.13$\pm$0.16} &  \textbf{9.68$\pm$0.44} &  \textbf{5.05$\pm$0.10} &  \textbf{13.14$\pm$0.28} &  \textbf{2.96$\pm$0.06} &  \textbf{6.55$\pm$0.26}
      \\ \cline{2-10}
      & EndtoEnd & 66.56$\pm$0.05 &  63.46$\pm$0.22 &  3.52$\pm$0.31 &  18.80$\pm$0.39 &  6.79$\pm$0.19 &  24.58$\pm$0.29 &  3.90$\pm$0.09 &  11.72$\pm$0.06 \\ \cline{2-10}
      & EWC & 64.75$\pm$0.14 &  65.70$\pm$0.37 &  1.36$\pm$0.41 &  24.18$\pm$0.23 &  9.68$\pm$0.31 &  18.10$\pm$0.69 &  5.98$\pm$0.07 &  7.71$\pm$0.15 \\ \cline{2-10}
      & Baseline & 62.64$\pm$0.26 &  64.40$\pm$0.29 &  0.21$\pm$0.32 &  25.01$\pm$0.33 &  10.17$\pm$0.36 &  18.80$\pm$0.31 &  6.49$\pm$0.10 &  8.42$\pm$0.11 \\ \hline
      \hline

    \parbox[t]{2mm}{\multirow{4}{*}{\rotatebox[origin=c]{90}{Best Cfg}}} 
      & \ours & \textbf{83.80} &  \textbf{84.28} &  \textbf{0.48} &  \textbf{2.56} &  \textbf{2.14} &  \textbf{12.11} &  \textbf{1.15} &  \textbf{4.57} \\ \cline{2-10}
      & EndtoEnd & 82.16 &  82.34 &  0.75 &  5.45 &  2.27 &  16.09 &  1.18 &  6.19 \\ \cline{2-10}
      & EWC & 72.70 &  72.44 &  2.46 &  22.65 &  7.35 &  18.83 &  4.11 &  6.62 \\ \cline{2-10}
      & Baseline & 70.24 &  71.06 &  1.48 &  25.45 &  8.18 &  19.37 &  4.72 &  7.14 \\ \hline 
      \hline
      
    \parbox[t]{2mm}{\multirow{4}{*}{\rotatebox[origin=c]{90}{Worst Cfg}}} 
      & \ours & \textbf{63.33} &  \textbf{65.21} &  \textbf{0.16} &  \textbf{5.11} &  \textbf{5.12} &  \textbf{10.54} &  \textbf{3.79} &  \textbf{6.92}
 \\ \cline{2-10}
      & EndtoEnd & 55.86 &  57.44 &  0.43 &  14.39 &  7.40 &  14.80 &  5.57 &  8.89 \\ \cline{2-10}
      & EWC & 50.52 &  49.85 &  2.51 &  24.55 &  10.41 &  19.18 &  8.33 &  9.38 \\ \cline{2-10}
      & Baseline & 44.72 &  46.25 &  0.58 &  29.43 &  12.19 &  22.73 &  10.75 &  11.06 \\ \hline
      
\end{tabular}}
\caption{\textbf{Incremental Learning Performance.} `Rand. Cfg' refers to random task configurations, (averaged over three random configurations with a standard deviation) `Best Cfg' refers to the best first task configuration (See text in Sec.\ref{sec:Best-config}), `Worst Cfg' refers to the worst first task configuration (See text in Sec.\ref{sec:worst-config}). Results in both simulated scenario are the averaged performance over multiple initialization.
EndToEnd refers to~\cite{castro2018eccv} and EWC refers to \cite{Kirkpatrick2017OvercomingCF}. 
Best results are in bold for each of the eight metrics. }
\label{table:alpha_curi}
\end{table*}

\begin{figure*}[t!]
\centering
\includegraphics[width=0.32\textwidth]{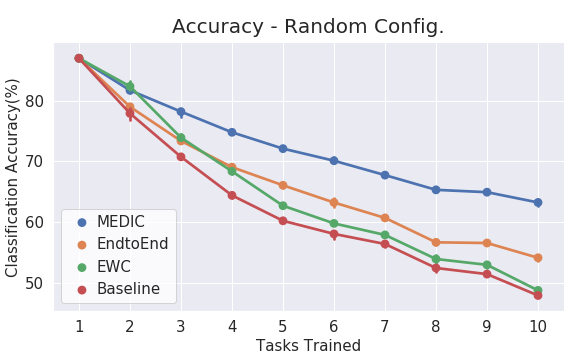}
\includegraphics[width=0.32\textwidth]{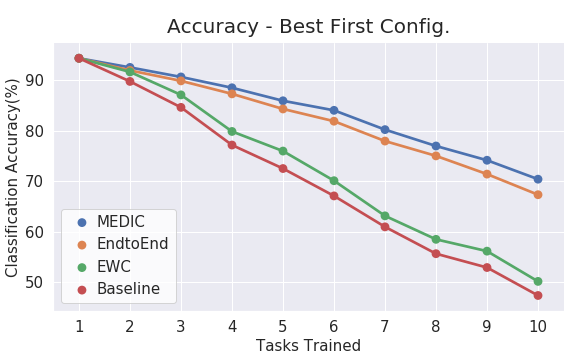}
\includegraphics[width=0.32\textwidth]{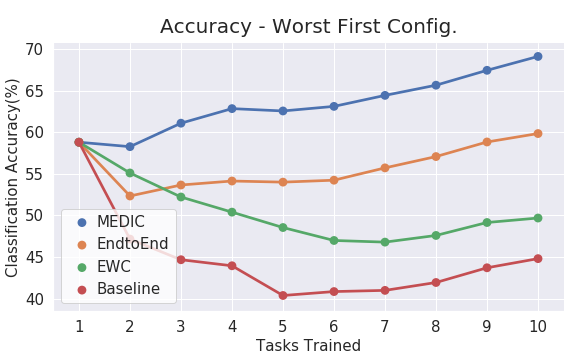}\\
\vspace{-.5em}
{\small (a)\hspace{17em}(b)\hspace{17em}(c)}
\vspace{-.5em}
\caption{\textbf{Classification Accuracy for Various Task Configurations.} (a) Random, (b) Best First, (c) Worst First. 
Plot (a) is shown with error bars from 3 trials. Refer to Section \ref{sec:sota} for details.}
\vspace{-1em}
\label{fig:alpha_curr}
\end{figure*}

We compare our model with the state-of-the-art (SOTA) models along with the baseline. 
The SOTA models include End to End Incremental Learning (EndToEnd) \cite{castro2018eccv} and our extension of Elastic Weight Consolidation with Sample Memeory (EWC), and summarize results in the top most group in the Table \ref{table:alpha_curi}.



The proposed \ours\ outperforms other methods in all metrics by large margins.
Note that the general trend in the F10 and the I10 measure is different from that of SDF10 and SDI10. 
The EndtoEnd shows worse F10 but better I10 performance when compared to the Baseline. 
It is counter-intuitive since the EndtoEnd uses distillation to overcome the forgetting while the Baseline does not.  
On the other hand, the proposed SDF10 and SDI10 metrics show more reasonable trends where the EndtoEnd shows better forgetting (SDF10) but worse intransigence (SDI10) compared to the Baseline. 

In addition, it is also counter-intuitive that the F10 of the Baseline is better than that of EWC's or EndtoEnd's. For possible explanations, refer to our argument in Sec.~\ref{sec:newmetric}.
In contrast, the proposed SDI10 and SDF10 metrics provide reasonable empirical values that capture the forgetting and intransigience.

\subsubsection{Simulated Task Configurations}
\label{sec:taskcfg}

\paragraph{Best First Configuration.}\label{sec:Best-config} 
As the best classes in terms of classification accuracy are given first, all methods show improved performance and the trends between the \ours\ and other methods are all preserved (the second group in Table \ref{table:alpha_curi}) with the improvement decreased. 
It is also expected since the uncertainty of the learned knowledge is small due to the easier classes being given first thus the benefits of the MER and the DOS decrease correspondingly.

Note that the forgetting (SDF10, $\text{SDI}_{avg}$) of EndToEnd is similar to \ours, suggesting that the MER is not much beneficial for forgetting in this configuration. Nevertheless, the difference in intransigence (SDI10, $\text{SDI}_{avg}$) between \ours\ and EndToEnd is still noticeable, which implies that the use of the MER and the DOS helps learning new classes (less intransigence).
Thus, even though the EndToEnd and the \ours\ show similar forgetting performance, the difference in intransigence allows \ours\ to outperform EndtoEnd by 1.6\% in the accuracy.

Still, the counter-intuitive results on the F10 is present again; the F10 measure is lower on the Baseline than it is on the EWC while the trends are more reasonable with SDF10.

\paragraph{Worst First Configuration.}\label{sec:worst-config} 
As the worst classes are given first, all methods show significant performance drops but the trends between \ours\ and other methods are preserved again (the third group in Table \ref{table:alpha_curi}) with the largest gain by \ours\ over the other methods among the three configurations in more than half of the metrics (Accuracy, A10, I10, SDF10, and $\text{SDF}_{avg}$). It implies that the benefit of the MER and the DOS is the most prominent in configuration where the teacher's soft labels are more uncertain from the beginning. 

However, the counter intuitive results on F10 continue; the Baseline achieves the best F10 in this configuration out of the three configurations. In contrast, the proposed SDF10 results in more reasonable values where the Baseline's forgetting being the worst out of all in the worst first configuration.

Unlike the other configurations, it is interesting to note that the accuracy at each task increases as the task progresses in the worst first configuration (compare Figure \ref{fig:alpha_curr}-(a), \ref{fig:alpha_curr}-(b) to Figure \ref{fig:alpha_curr}-(c)).

\begin{table*}[!th]
\centering
\resizebox{0.77\paperwidth}{!}{
\begin{tabular}{c|c|c|c|c||c|c|c|c}
\hline
Methods & Accuracy $\uparrow$ & A10 $\uparrow$ & F10 $\downarrow$ & I10 $\downarrow$ & SDF10 $\downarrow$ & SDI10 $\downarrow$ & $\text{SDF}_{avg}$ $\downarrow$ & $\text{SDI}_{avg}$ $\downarrow$ \\ \hline
\ours & \textbf{69.54$\pm$0.44} & \textbf{69.56$\pm$0.69} &  0.41$\pm$0.13 &  \textbf{13.78$\pm$0.80} &  \textbf{5.68$\pm$0.43} &  \textbf{17.61$\pm$0.35} &  \textbf{3.38$\pm$0.30} &  \textbf{8.72$\pm$0.23} \\ \hline
EndtoEnd & 62.73$\pm$0.15 &  61.49$\pm$0.26 &  1.46$\pm$0.37 &  22.54$\pm$0.33 &  8.03$\pm$0.40 &  24.70$\pm$0.94 &  5.16$\pm$0.07 &  11.49$\pm$0.06 \\ \hline
EWC & 56.63$\pm$0.23 &  57.95$\pm$1.30 &  1.93$\pm$1.75 &  34.78$\pm$0.24 &  13.01$\pm$0.21 &  24.05$\pm$1.75 &  8.65$\pm$0.07 &  9.98$\pm$0.07 \\ \hline
Baseline & 52.49$\pm$0.17 &  56.79$\pm$0.39 &  \textbf{0.05$\pm$0.21} &  38.40$\pm$0.58 &  14.32$\pm$0.36 &  24.03$\pm$0.57 &  9.74$\pm$0.08 &  10.85$\pm$0.17 \\ \hline
\end{tabular}}
\caption{\textbf{Performance on CIFAR100 with Reduced Budget of Old Class Samples (2,000 to 1,000).} Best results are in bold across the methods.}
\label{table:budget1000}
\end{table*}

\begin{figure}[t]
\includegraphics[width =\columnwidth]{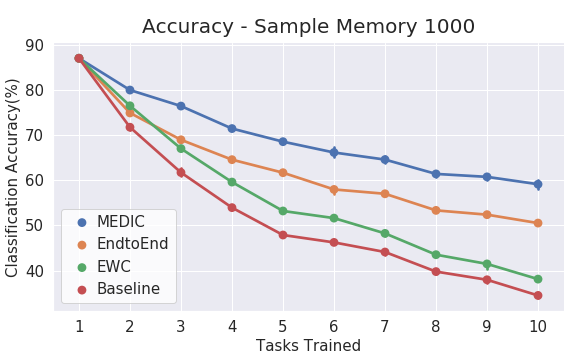}
\vspace{-8mm}
\caption{\textbf{Smaller Budget of Sample Memory (1,000).} 
Plots are shown with error bars from 3 trials. Refer to Section \ref{sec:budget1000} for details.}
\vspace{-1em}
\label{fig:budget1000}
\end{figure}

\begin{table*}[!t]
\centering
\resizebox{0.77\paperwidth}{!}{
\begin{tabular}{c|c|c|c|c||c|c|c|c}
\hline
Methods & Accuracy $\uparrow$ & A10 $\uparrow$ & F10 $\downarrow$ & I10 $\downarrow$ & SDF10 $\downarrow$ & SDI10 $\downarrow$ & $\text{SDF}_{avg}$ $\downarrow$ & $\text{SDI}_{avg}$ $\downarrow$ \\ \hline
\ours & \textbf{49.88$\pm$0.38} &  \textbf{45.97$\pm$0.77} &  4.54$\pm$0.75 &  \textbf{14.83$\pm$1.37} &  \textbf{9.24$\pm$0.92} &  \textbf{28.39$\pm$10.48} &  \textbf{5.37$\pm$0.24} &  12.90$\pm$4.31 \\ \hline
EndtoEnd & 45.55$\pm$0.76 &  45.58$\pm$1.53 &  \textbf{1.22$\pm$1.52} &  20.09$\pm$2.13 &  10.73$\pm$2.08 &  30.64$\pm$9.96 &  6.37$\pm$1.12 &  13.23$\pm$3.21 \\ \hline
EWC & 43.92$\pm$1.17 &  41.79$\pm$1.12 &  3.84$\pm$0.87 &  23.31$\pm$1.08 &  12.61$\pm$1.63 &  33.55$\pm$7.19 &  8.73$\pm$0.16 &  \textbf{12.57$\pm$2.69} \\ \hline
Baseline & 42.37$\pm$0.06 &  43.23$\pm$0.07 &  1.55$\pm$0.29 &  21.27$\pm$1.14 &  12.91$\pm$1.24 &  31.28$\pm$10.80 &  9.98$\pm$1.45 &  13.08$\pm$4.54 \\ \hline
\end{tabular}}
\caption{\textbf{Performance on TinyImagenet Dataset.} Best results are in bold in the comparing methods.}
\label{table:timage}
\end{table*}

\subsection{Reduced Memory Budget}
\label{sec:budget1000}
For a more cost efficient incremental set-up, we reduce the memory size for the old classes from 2,000 to 1,000 on the CIFAR100 dataset, and repeat the same experiments on the random task configuration. We summarize the results in Table \ref{table:budget1000}.
While all methods suffer from performance drop in all metrics, the \ours\ suffers the least with the minimum drop in performance, thus resulting in a larger gap between the \ours\ and other methods compared to the first group in Table~\ref{table:alpha_curi}. 
It is because the soft labels get even more uncertain with a smaller budget, leading to larger gains when using the MER and the DOS. 
Further, the smaller budget worsens the class imbalance, resulting in more benefit by the DOS to allieviate the class imbalance.

Another interesting observation is that the SDI10 in the worst first configuration is generally lower or similar to the SDI10 measure in the best first configuration, but the $\text{SDI}_{avg}$ does not. One possible explanation is that the SDI10 is more heavily affected by the last classes that were added, which are the easiest classes in the worst first configuration. 
Correspondingly, the $SDI10$ is lower than $SDI_{avg}$. 

However, ${SDI}_{avg}$ measures intransigence at each incremental step, hence the harder classes that were given first have a bigger impact on it, resulting in higher figure. Hence, using both the $SDF10$, $SDI10$, ${SDI}_{avg}$ and ${SDI}_{avg}$ measures will provide a more thorough evaluation of incremental learning models on different task configurations.

\subsection{Experiments on TinyImageNet}
\label{sec:timage}
We also conduct experiments on the TinyImageNet dataset and summarize the results in Table \ref{table:timage}. 
The \ours\ shows the best forgetting and intransigence, leading to a noticable gain in classification accuracy compared to other methods. 

Counter-intuitive results on the F10 continue; the F10 of the baseline is lower than the F10 of EWC while on the SDI and SDF, the baseline performs the worst. 
The mixed result in F10 is most likely due to two reasons; first, it is due to the lower overall performance in TinyImageNet, leading to reduced differences in task accuracies. Second, the 200 classes in TinyImageNet are more likely to be confused, thus leading the F10 to interpret simple confusion with confusable classes as forgetting. 


\section{Conclusion}
We propose a method called \ours\ for incremetal learning with less forgetting \emph{and} less intransigence by a regularized knowledge distillation loss with the Maximum Entropy Regularization (MER) and the \emph{DropOut Sampling (DOS)}.
We also propose new metrics for the forgetting and the intransigence by tracking how a sample's confusion behaves at the task transition, called \emph{Sample Dynamics-Forgetting (SDF)} and \emph{Sample Dynamics-intransigence (SDI)}.

We show that the MER and the DOS significantly improve performance in terms of the final classification accuracy, the F and the I measure (for forgetting and intransigence, respectively) proposed by~\cite{Chaudhry_2018_ECCV}, and our proposed SDI and SDF in an ablation study on CIFAR100.
We also show that the \ours\ significantly outperforms the state-of-the-art incremental learning methods including elastic weight consolidation (EWC)~\cite{Kirkpatrick2017OvercomingCF} and End-to-End incremental learning method~\cite{castro2018eccv} on two datasets (CIFAR100 and TinyImageNet).
We further analyze the methods in two extreme scenarios of incremental learning task configurations, the best-first and the worst-first, and show that the proposed method outperforms other methods in all scenarios.
We additionally show that the proposed method is particularly useful in a budgeted incremental setup where the budget for the memory of old classes' samples is small (1,000) by outperforming other incremental learning methods with larger margin than in a setup with a higher budget (2,000).




\bibliographystyle{icml2018}

\end{document}